\documentclass[sigconf,screen]{acmart}

\usepackage{multirow}
\usepackage{booktabs}

\renewcommand\footnotetextcopyrightpermission[1]{}
\acmConference[arXiv]{arXiv preprint}{2026}{}

\begin{document}

\title{State Beyond Appearance: Diagnosing and Improving State Consistency in Dial-Based Measurement Reading}


\author{Yuanze Hu}
\affiliation{%
  \institution{Beijing Advanced Innovation Center for Future Blockchain and Privacy Computing,\\ Beihang University }
  \city{Beijing}
  \country{China}}
\email{huyuanze@buaa.edu.cn}
\author{Gen Li}
\affiliation{%
  \institution{Beijing Advanced Innovation Center for Future Blockchain and Privacy Computing,\\ Beihang University }
  \city{Beijing}
  \country{China}}
\author{Yuqin Lan}
\affiliation{%
  \institution{Beijing Advanced Innovation Center for Future Blockchain and Privacy Computing,\\ Beihang University }
  \city{Beijing}
  \country{China}}
\author{Qingchen Yu}
\affiliation{%
  \institution{Beijing Advanced Innovation Center for Future Blockchain and Privacy Computing,\\ Beihang University }
  \city{Beijing}
  \country{China}}
\author{Zhichao Yang}
\affiliation{%
  \institution{Beihang University }
  \city{Beijing}
  \country{China}}
\author{Junwei Jing}
\affiliation{%
  \institution{Fudan University}
  \city{Shanghai}
  \country{China}}
\author{Zhaoxin Fan}
\affiliation{%
  \institution{Beijing Advanced Innovation Center for Future Blockchain and Privacy Computing,\\ Beihang University }
  \city{Beijing}
  \country{China}}
\author{Xiaotie Deng}
\affiliation{%
  \institution{Peking University}
  \city{Beijing}
  \country{China}}

\renewcommand{\shortauthors}{Hu et al.}

\begin{abstract}
Multimodal large language models (MLLMs) have achieved impressive progress on general multimodal tasks, yet they remain brittle on dial-based measurement reading. In this paper, we study this problem through controlled benchmarks and feature-space probing, and show that current MLLMs not only achieve unsatisfactory accuracy on dial-based readout, but also suffer sharp performance drops under viewpoint and illumination changes even when the underlying dial state remains fixed. Our probing analysis further reveals that same-state samples under appearance variation are not consistently clustered, while neighboring states fail to preserve the local structure implied by continuous dial values. These findings suggest that existing MLLMs largely ignore the intrinsic state geometry of dial measurement tasks and instead rely on superficial appearance cues. Motivated by this diagnosis, we propose TriSCA, a tri-level state-consistent alignment framework for dial-based measurement reading. Specifically, TriSCA consists of state-distance-aware representation alignment, metadata-grounded observation-to-state supervision, and state-aware objective alignment. Extensive ablation studies and evaluation experiments on controlled clock and gauge benchmarks, together with evaluation on an external real-world benchmark, demonstrate the effectiveness of our method.
\end{abstract}



\keywords{Multimodal Large Language Models, State Consistency, Dial-Based Readout, Fine-Grained Visual Understanding}

\maketitle

\section{Introduction}
\label{sec:intro}

\begin{figure}[t]
  \centering
  \includegraphics[width=0.93\linewidth]{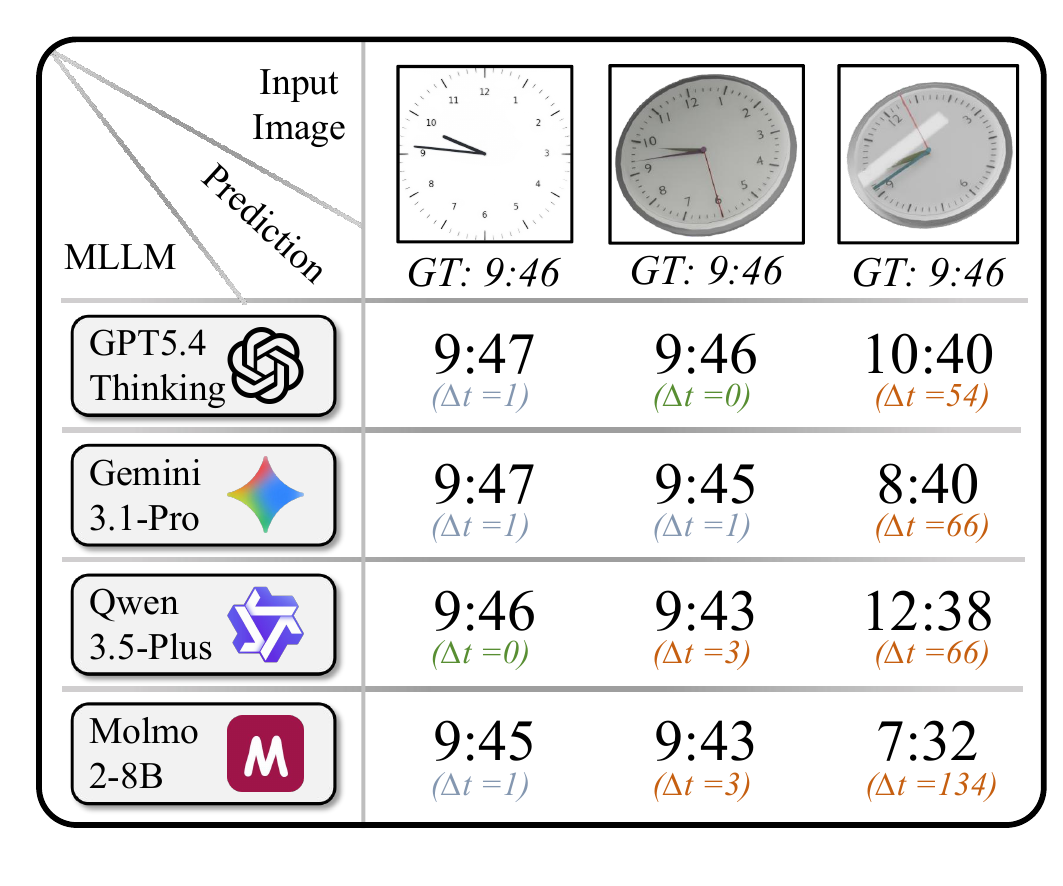}
\caption{%
  Current MLLMs are fragile under appearance shifts. Performance drops sharply as viewpoint and lighting deviate from canonical conditions. 
}
  \label{fig:phenomenon}
\end{figure}

Multimodal large language models (MLLMs) have achieved strong performance across a broad range of vision-language tasks, yet this aggregate success conceals a persistent weakness in fine-grained visual perception~\cite{comanici2025gemini, yue2024mmmu,yue-etal-2025-mmmu,singh2019towards, wang2024qwen2, deitke2025molmo,vicuna, cogvlm}. Recent targeted evaluations show that when a task demands precise perceptual grounding rather than high-level semantic recognition, state-of-the-art MLLMs fall dramatically short of human performance~\cite{fu2024blink, kamoi2024visonlyqa, kanade2025doyouseeme, stogiannidis2025mindthegap,moe-llava}. A plausible explanation is that current MLLMs often fail to recover the fine-grained visual evidence a task demands, even when they can still produce fluent, semantically coherent responses. This gap becomes most consequential in tasks where the answer is not a semantic category, but an exact physical state that must be read from visual evidence~\cite{hall_tune, minigpt4,sharegpt4v,dola}.

Dial-based measurement reading is precisely such a task. When a model reads an analog clock, interprets a pointer gauge, or processes an industrial dial meter, it must map low-level geometric evidence---pointer orientation, angular displacement, and scale alignment---to a continuous or finely discretized physical state~\cite{salomon2022image, leonalcazar2024gauges}. This is fundamentally different from category recognition: the label space is intrinsically ordered and structured, and a representation well aligned with the task should preserve meaningful neighborhood relations among nearby states. At the same time, the same physical state can project onto visually distinct images under viewpoint shifts and illumination changes, demanding a degree of appearance invariance that standard visual pretraining does not reliably guarantee~\cite{salomon2022image, reitsma2024under,hill}. MeasureBench formalizes this broader setting and confirms that even frontier VLMs struggle substantially, identifying indicator localization and fine-grained spatial grounding as central failure modes~\cite{lin2025measurebench}. The practical urgency is well established: automatic meter reading, robotic inspection, and safety-critical field monitoring all depend on reliable state recovery under unconstrained imaging conditions~\cite{salomon2022image, reitsma2024under}.

\begin{figure}[t]
  \centering

  \includegraphics[width=0.30\linewidth]{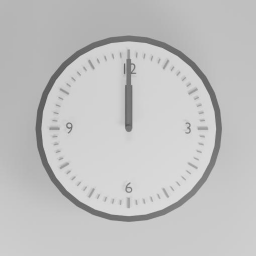}\hfill
  \includegraphics[width=0.30\linewidth]{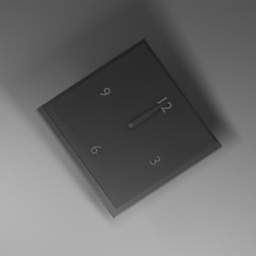}\hfill
  \includegraphics[width=0.30\linewidth]{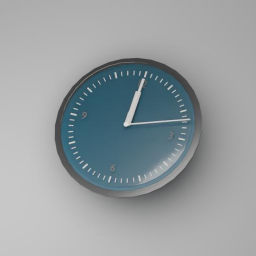}

  \vspace{2pt}
  {\small (a) Same state under different appearance conditions.}

  \vspace{5pt}

  \includegraphics[width=0.30\linewidth]{img/fig2_same1.pdf}\hfill
  \includegraphics[width=0.30\linewidth]{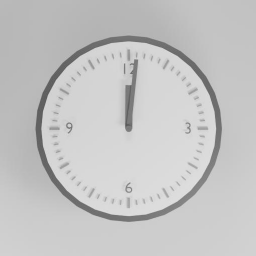}\hfill
  \includegraphics[width=0.30\linewidth]{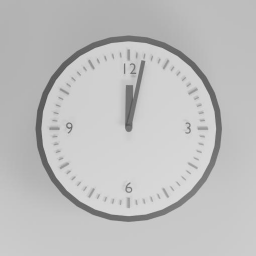}

  \vspace{2pt}
  {\small (b) Neighboring states with subtle visual differences.}

\caption{%
  \textbf{Dial-based readout is structured by state relations, not isolated labels.}
  (a) The same underlying state can appear very different under viewpoint or illumination changes, requiring invariance to appearance-only perturbations.
  (b) Neighboring dial states may look highly similar yet still demand precise discrimination.
  Reliable readout therefore requires both cross-condition consistency and sensitivity to subtle state changes.
}
  \label{fig:task_structure}
\end{figure}

Recent work confirms that time understanding and visual measurement reading remain far from solved for current MLLMs~\cite{saxena2025lost, lin2025measurebench}, and the community has recognized clock reading as a capability worth targeting directly~\cite{deitke2025molmo}. Yet existing discussions often frame the difficulty either as a data-coverage problem or as an isolated benchmark failure. What they do not address is the deeper question of whether the model has learned a representation organized by physical state at all. If visual features are structured primarily by appearance---lighting, texture, or viewpoint---rather than by the underlying state they encode, then performance will become fragile under appearance shifts regardless of whether the state has changed, and subtle state changes may remain invisible whenever their visual footprint is small. We argue that this representational misalignment, not merely low accuracy, is the key problem in dial-based measurement reading. Testing this hypothesis requires a controlled setting that isolates appearance variation while holding the underlying state fixed, which existing broad evaluations do not directly provide~\cite{lin2025measurebench}.

To test this hypothesis, we conduct controlled experiments and feature-space probing using matched same-state and nearby-state sample pairs. Our findings confirm the concern. First, even strong open MLLMs achieve unsatisfactory accuracy on dial-based reading and degrade sharply under viewpoint tilt and illumination changes---despite the underlying state remaining fixed. Second, probing the visual feature space reveals two complementary deficiencies: same-state samples across appearance conditions are not consistently clustered (poor cross-view compactness), and neighboring states fail to preserve the local ordering implied by continuous dial values (insufficient fine-grained state separability). Together, these findings characterize the central failure as a lack of \emph{state consistency}---the simultaneous inability to remain invariant under state-preserving appearance variation and to remain sensitive to subtle state-changing differences. This interpretation is consistent with broader analyses that trace MLLM failures to unstable fine-grained representations and deficient spatial grounding~\cite{fu2024blink, stogiannidis2025mindthegap}, and suggests that the learned feature space in current MLLMs tends to be organized more strongly by appearance than by physical state.

Motivated by this diagnosis, we propose \textbf{TriSCA}, a tri-level state-consistent alignment framework that improves dial-based measurement reading from three complementary levels: representation, reasoning, and objective alignment. At the representation level, we introduce \emph{state-distance-aware representation alignment}, which organizes the feature space according to continuous state proximity, enforcing tighter cross-condition intra-state compactness and sharper inter-state boundaries calibrated to physical state difference. At the reasoning level, we construct \emph{metadata-grounded observation-to-state supervision} that explicitly guides the model through a grounded readout process---identifying the relevant indicator, estimating its dial-relative position, and mapping it to the calibrated state rather than collapsing the task into a single holistic prediction. At the objective level, we apply \emph{state-aware objective alignment}, replacing sparse exact-match rewards with continuous state-distance-based reward signals that align model updates with the inherent geometry of the dial state space.

\begin{figure*}[t]
  \centering
  \includegraphics[width=0.8\textwidth]{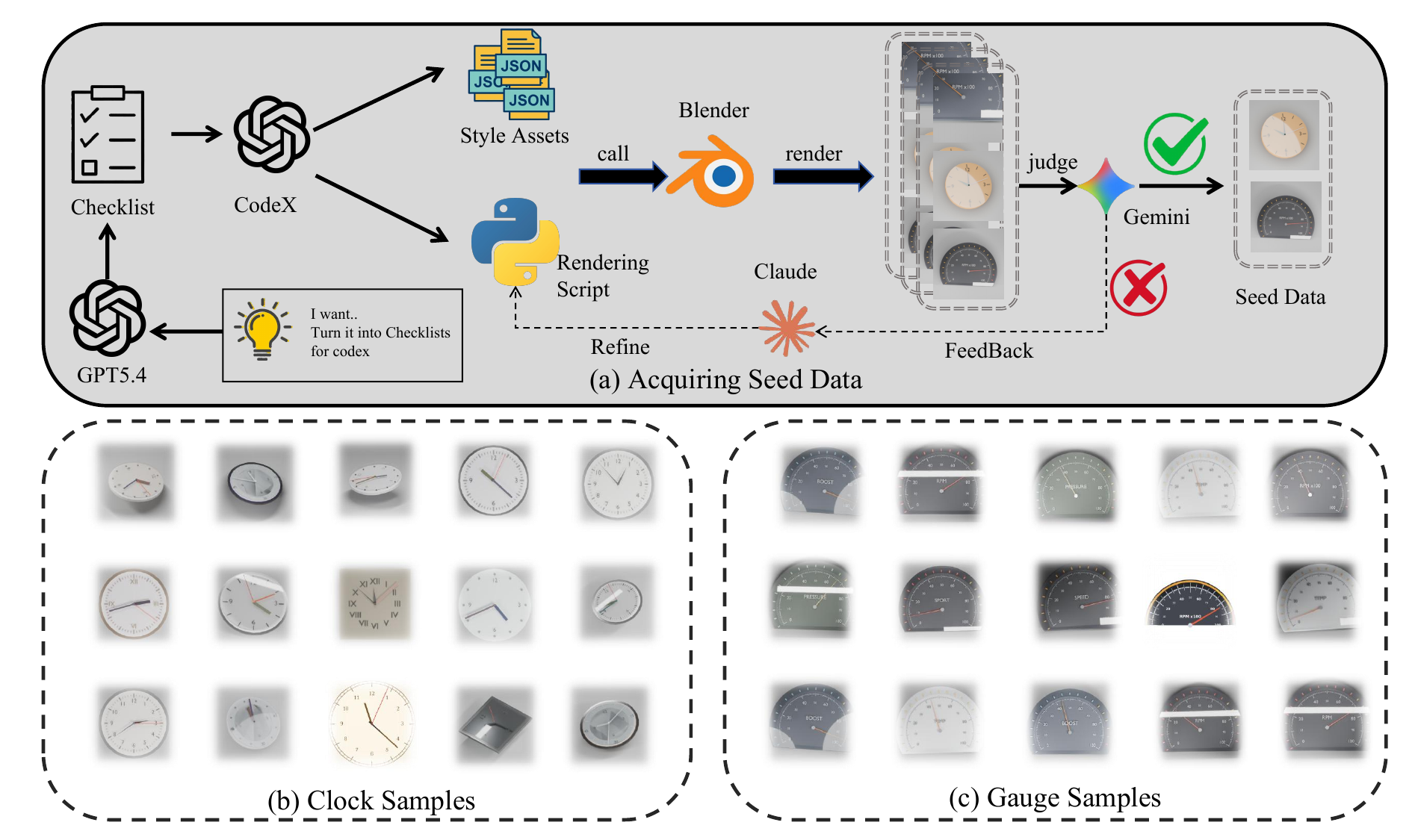}
\caption{%
    \textbf{Controlled synthesis enables state-aware diagnosis and training for both clocks and gauges.}
    (a) Our synthesis pipeline acquires seed dial data and iteratively builds qualified synthetic assets through design, code execution, rendering, and feedback-driven refinement.
    (b) Example synthesized clock samples.
    (c) Example synthesized gauge samples.
    This design decouples underlying state from appearance variation and provides fine-grained neighboring-state samples for both probe-based analysis and subsequent multi-stage training.
}
  \label{fig:synth_pipeline}
\end{figure*}

In summary, this paper makes three contributions. (1) We formulate \emph{state consistency} as a key missing capability in dial-based measurement reading and introduce controlled benchmarks that isolate viewpoint and illumination perturbations while holding the underlying dial state fixed. (2) We provide feature-space probing evidence that current MLLMs encode appearance more strongly than physical state in this setting, yielding a concrete representational diagnosis of their failures. (3) We propose a unified tri-level training framework---spanning state-distance-aware representation alignment, metadata-grounded observation-to-state supervision, and state-aware objective alignment---and demonstrate consistent gains in robustness under viewpoint and illumination changes, together with improved transfer to external real-world evaluation.

\section{Related Work}
\label{sec:related}

\subsection{Fine-Grained Visual Perception, Visual Measurement Reading, and Analog Instrument Understanding}

A growing body of work has shown that strong aggregate benchmark performance does not imply reliable fine-grained visual perception in MLLMs. MMStar argues that many existing evaluations overestimate multimodal competence because they admit language shortcuts or data leakage, while BLINK demonstrates that state-of-the-art MLLMs remain far below human performance on core perceptual tasks such as relative depth estimation, visual correspondence, and multi-view reasoning~\cite{ascd,shikra, fu2024blink, yang2022s, liu2024ocrbench,cheng2025v}. More targeted studies further show that current LVLMs struggle to perceive basic geometric information and spatial relations even when such judgments are nearly trivial for humans~\cite{kamoi2024visonlyqa, kanade2025doyouseeme, stogiannidis2025mindthegap}. Together, these works suggest that modern MLLMs can often produce plausible responses without fully recovering the fine-grained visual evidence required by perception-intensive tasks.

A particularly relevant instance of this broader limitation is \emph{visual measurement reading}, where the model must recover an exact value from image evidence rather than recognize a semantic category. MeasureBench formalizes this setting and shows that even frontier VLMs struggle substantially, identifying indicator localization and fine-grained spatial grounding as central bottlenecks~\cite{lin2025measurebench}. A parallel line of work in traditional computer vision has studied analog instrument reading through specialized pipelines for dial meters, pointer gauges, and related devices, often relying on geometric rectification, pointer detection, synthetic rendering, or task-specific data augmentation~\cite{salomon2020deep, salomon2022image, param_hall, param_hall1,leonalcazar2024gauges, reitsma2024under}. Prior work has also addressed analog clock reading explicitly through synthetic-data training and sim-to-real transfer~\cite{yang2022s}. These studies establish both the practical importance and the technical difficulty of robust reading under unconstrained conditions such as oblique viewpoints, variable illumination, reflections, and occlusion, but they do not address how general-purpose MLLMs represent the underlying state structure of such tasks.

\subsection{Clock and Dial Reading in MLLMs}

Within the broader measurement-reading literature, clock and dial reading has emerged as a particularly revealing testbed for MLLMs. Lost in Time shows that analogue clock and calendar understanding remain far from solved for current multimodal models~\cite{saxena2025lost}. Molmo further recognizes clock reading as a capability worth targeting explicitly and introduces PixMo-Clocks to strengthen this skill through dedicated synthetic supervision~\cite{deitke2025molmo}. More recent work asks an even sharper question: whether MLLMs have truly learned to tell time on analog clocks or merely fit superficial visual patterns correlated with specific hand configurations~\cite{fu2025have}. Very recent efforts also begin to improve analog clock reading in real-world environments with targeted datasets and task-specific alignment, especially under clutter, occlusion, and lighting variation~\cite{choi2026s}. These studies establish that clock and dial reading are persistent weaknesses of current MLLMs, and that targeted data or task-specific fine-tuning can improve performance. However, they still primarily treat the problem as a capability gap or a data-coverage issue.

Our work differs in focus. Rather than treating dial reading only as a task-specific weakness, we diagnose it as a representation problem: whether the learned feature space is organized by physical state and remains stable under state-preserving appearance changes. This perspective is informed by prior work on structured representation learning, grounded intermediate supervision, and reinforcement-based alignment, which motivate our design without directly addressing the continuous-state geometry of dial-based measurement reading~\cite{schroff2015facenet, khosla2020supervised, zhang2023multimodal, cheng2025visualthoughts, he2026finer, shao2024deepseekmath}. Our contribution is therefore not any single ingredient in isolation, but the task-specific unification of these ideas toward improving \emph{state consistency} in dial-based reading.

\section{Diagnostic Setting and Representation Diagnosis}
\label{sec:diagnosis}

\paragraph{Unified Dial-State Formulation.}
We view dial-based readout as recovering a latent physical state from visual evidence under appearance variation. Each sample is factorized into a dial state $s$ and an appearance condition $a$. The state $s$ captures the underlying physical quantity encoded by the dial, while $a$ collects nuisance factors such as viewpoint, illumination, reflection, and blur. For clocks, $s$ is represented by the underlying time state or its induced hand-angle configuration. For gauges, $s$ is represented by a calibrated scalar value together with its corresponding pointer angle under the dial geometry. A reliable model should therefore satisfy two requirements simultaneously: invariance to changes in $a$ when $s$ is fixed, and sensitivity to small but meaningful changes in $s$ under comparable appearance.
\begin{figure*}[t]
\centering
\includegraphics[width=0.8\textwidth]{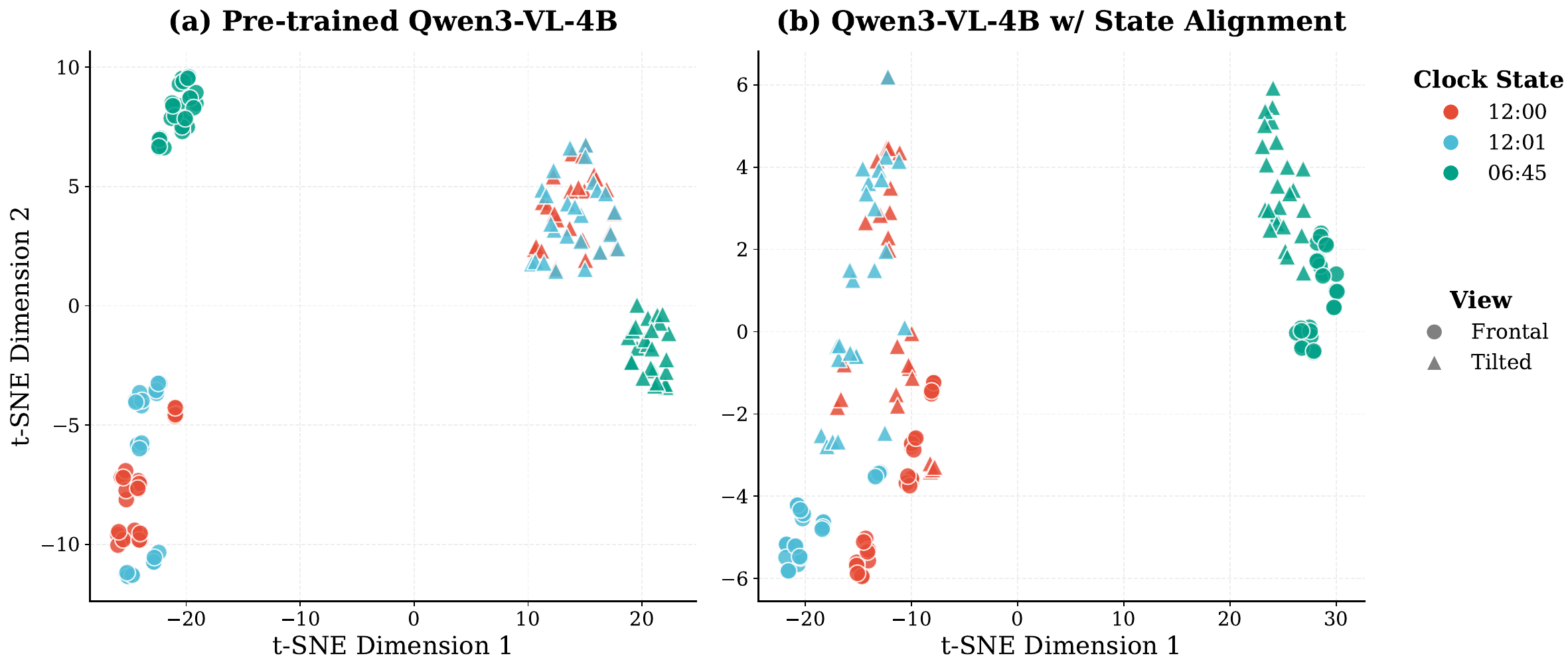}
\caption{%
\textbf{State-aware alignment reshapes the feature space toward dial-state consistency.}
(a) Off-the-shelf visual features are not well organized by the fine-grained state structure required for dial-based reading: same-state samples are insufficiently compact across appearance conditions, and neighboring states such as 12{:}00 and 12{:}01 remain weakly distinguishable.
(b) After representation alignment, the feature space exhibits improved cross-condition compactness and clearer local separation between neighboring states.
These results indicate that our method improves both state-preserving invariance and fine-grained state discriminability.
}
\label{fig:tsne_compare}
\end{figure*}
\subsection{Task Observation and Structural Motivation}
\label{subsec:observation}

As shown in Figure~\ref{fig:phenomenon}, current MLLMs exhibit a systematic performance drop on dial-based reading as appearance deviates from the canonical condition. As shown in Figure 1, state-preserving appearance shifts (e.g., viewpoint tilt, illumination) alone cause substantial prediction drift. This phenomenon suggests that the difficulty of dial-based reading is not merely low accuracy on a challenging benchmark, but a more fundamental failure to maintain stable state estimation under appearance change.

A key reason is that dial-based reading is structurally different from standard category recognition. The target is not a semantic label but a physical state encoded by geometric evidence such as pointer orientation, angular displacement, and scale alignment. As illustrated in Figure~\ref{fig:task_structure}, two kinds of relations are intrinsic to this task: the same state can correspond to visually different images under viewpoint or lighting changes, while neighboring states can appear highly similar yet still require precise discrimination. A reliable model must therefore remain stable under appearance-only perturbations while staying sensitive to subtle state differences.

This observation motivates our central hypothesis: dial-based measurement reading should not be treated as a collection of isolated labels but modeled as a structured prediction problem whose samples are linked by continuous state relations. A model that ignores this structure and organizes its representation primarily by superficial appearance will inevitably become brittle under state-preserving appearance changes and imprecise when distinguishing nearby states.

\subsection{From Clock Phenomenon to Broader Dial Tasks}
\label{subsec:clock_to_dial}

Our initial observations and analyses are conducted on analog clocks because clock reading provides a clean and intuitive testbed: the state is easy to define, neighboring states are naturally ordered, and the effects of viewpoint and illumination change can be clearly controlled. However, the issue we study is not specific to clocks. It reflects a broader challenge shared by dial-based measurement tasks, including pointer gauges and analog meters.

These tasks share the same underlying structure: the model must infer a target state from the geometric relation between an indicator and a calibrated dial. The visual appearance of the same state may change substantially under different imaging conditions, while nearby states often differ only in fine-grained local evidence. As a result, the same failure mode observed on clocks can also arise in other dial domains. Motivated by this broader perspective, we extend our study beyond clocks and build a gauge synthesis pipeline in addition to the clock pipeline, allowing us to test whether the observed fragility is a general property of dial-based reading rather than a clock-specific artifact.

\subsection{Controlled Data Synthesis Pipeline}
\label{subsec:synth_pipeline}

To study state consistency systematically, we require data in which the underlying dial state and the visual appearance can be manipulated in a controlled manner. Real-world collections alone are insufficient because viewpoint, illumination, dial style, and state often change together, making it difficult to isolate the source of model failure. We therefore construct controlled synthesis pipelines for both clocks and gauges, as illustrated in Figure~\ref{fig:synth_pipeline}.

Our synthesis pipeline factorizes each sample into an underlying dial state and a set of appearance variables. The state determines the calibrated dial reading, while appearance variables control nuisance factors such as camera pose, lighting condition, reflection strength, and blur. This factorization allows us to construct two key diagnostic relations in a controlled way: (1) \emph{same-state pairs}, where the dial state is fixed and only appearance changes, and (2) \emph{neighboring-state pairs}, where appearance is kept comparable while the dial state changes slightly.

For clocks, the state corresponds to the underlying time configuration. For gauges, the state corresponds to a calibrated scalar dial value together with its induced pointer angle under the dial geometry. In our gauge setting, the dial is modeled as a single-pointer half-circle instrument, where the scalar value is mapped to pointer angle through the gauge calibration. This shared state-centric view allows the same synthesis logic to extend naturally from clock readout to more general pointer-based dial instruments.

Appearance variation is introduced through controlled camera and rendering parameters rather than uncontrolled web-scale data mixing. We vary viewpoint, lighting environment, reflection/glare, and blur-related degradation while keeping the state fixed when constructing same-state pairs. The renderer records these factors as structured metadata, which supports benchmark split construction and difficulty bucketing. These synthesis pipelines are not merely sources of additional training data; more importantly, they provide the controlled relations needed for both diagnosis and alignment, including state-preserving appearance perturbations and fine-grained neighboring-state pairs used across all three training stages.

\subsection{Probe-Based Diagnosis of Representation Misalignment}
\label{subsec:probe}

\begin{table}[t]
\centering
\small
\setlength{\tabcolsep}{5pt}
\renewcommand{\arraystretch}{1.1}
\caption{
  \textbf{Quantitative evaluation of the feature space.}
  State-aware alignment improves exact same-state retrieval (Recall@1) and overall cluster compactness/separability (Silhouette Score), corroborating the improved cross-condition compactness and neighboring-state separability suggested by the feature visualization.
}
\label{tab:retrieval_metrics}
\begin{tabular}{lcc}
\toprule
\textbf{Model} & \textbf{R@1 (\%) $\uparrow$} & \textbf{Silhouette $\uparrow$} \\
\midrule
Qwen3-VL-4B-Instruct & 77.33 & 0.0725 \\
\textbf{Ours (Aligned)} & \textbf{90.00} & \textbf{0.3409} \\
\bottomrule
\end{tabular}
\end{table}

Based on the structural motivation above, we further ask whether current visual representations actually reflect the state structure required by dial-based reading. If they do, same-state samples should remain relatively stable across appearance changes, while neighboring states should preserve a locally meaningful organization that still supports precise readout. To examine this, we probe the feature space using controlled sample relations constructed from our synthesized data, comparing samples that share the same state but differ in appearance and samples that are close in state but differ in subtle pointer geometry~\cite{hu2025tinyalign}. Figure~\ref{fig:tsne_compare} visualizes the resulting feature distribution.

Two observations are particularly important. First, same-state samples under different appearance conditions are not consistently compact, indicating limited invariance to state-preserving perturbations. Second, neighboring states such as 12{:}00 and 12{:}01 remain insufficiently distinguishable, suggesting weak local discriminability for fine-grained state estimation. The issue is therefore not simply that features are poorly separated in a generic sense; rather, the learned geometry does not properly encode the state relations this task depends on: it is neither stable enough across appearance changes nor locally organized enough for precise readout.

This diagnosis motivates our method design. Since the base representation does not adequately respect the continuous structure of dial states, we improve dial-based reading through stage-wise alignment at the levels of representation, reasoning, and training objective.

\section{TriSCA: State-Consistent Alignment}
\label{sec:method}
\begin{figure}[t]
  \centering
  \includegraphics[width=0.8\linewidth]{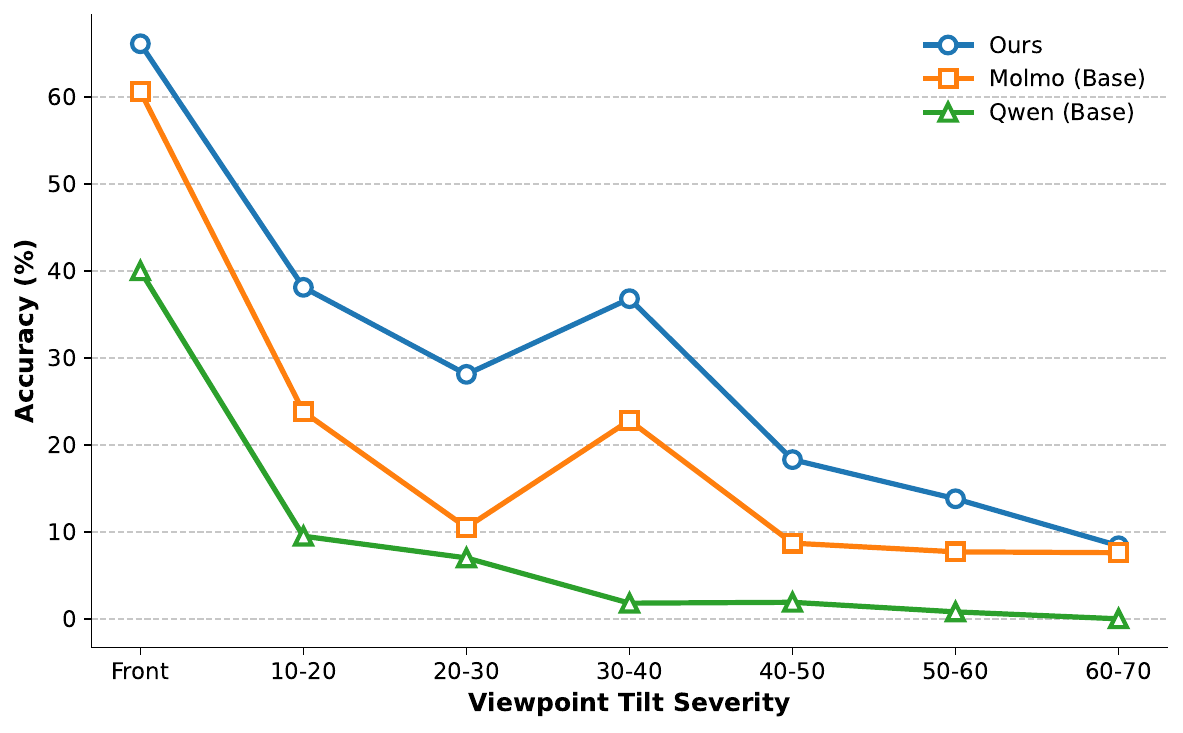}
\caption{%
  \textbf{State-consistent alignment slows performance degradation under increasing appearance shifts.}
  We bucket evaluation samples by perturbation difficulty and report accuracy as viewpoint or illumination severity increases.
  Aligned models degrade substantially more slowly, indicating improved robustness to state-preserving appearance changes.
}
  \label{fig:degradation_curve}
\end{figure}

\subsection{Stage 1: State-Distance-Aware Representation Alignment}
\label{subsec:stage1}

Stage~1 addresses the representational mismatch diagnosed in Section~\ref{sec:diagnosis}. Our goal is not merely to make dial features more separable in a generic sense, but to organize them according to physically meaningful state relations.

We first freeze the language model parameters and update only the visual tower, so that the alignment signal directly reshapes visual representations without disturbing the base decoder~\cite{bai2025qwen3}. For each training triplet $(a,p,n)$, the anchor $a$ is paired with answer supervision, while the positive $p$ and negative $n$ are used to construct a contrastive state-structured visual objective. The anchor and positive share the same underlying dial state but differ in appearance, whereas the negative is sampled from a different dial state with a controlled state gap.

Let $f(\cdot)$ denote the pooled visual representation of an input image. We compute a triplet-style loss
\[
\mathcal{L}_{\text{tri}}
=
\frac{1}{B}\sum_{i=1}^{B}
\left[
\|f(a_i)-f(p_i)\|_2
-
\|f(a_i)-f(n_i)\|_2
+
m_i
\right]_+,
\]
where $m_i$ is a sample-specific margin associated with the $i$-th triplet, assigned according to the state gap between the anchor and the negative so that pairs with larger state discrepancy are encouraged to remain farther apart in the learned feature space.

We combine this representation loss with standard answer supervision on the anchor branch:
\[
\mathcal{L}_{\text{stage1}}
=
\mathcal{L}_{\text{ce}}
+
\lambda_t \mathcal{L}_{\text{tri}},
\]
where $\lambda_t$ is gradually warmed up during training to stabilize optimization. For clocks, the state gap is computed from the temporal state or its induced hand-angle configuration; for gauges, it is computed from the dial value difference or the corresponding pointer-angle difference under gauge calibration. Stage~1 thus explicitly enforces both cross-condition compactness for same-state samples and state-aware separation for different dial readings.
\begin{table*}[t]
\centering
\small
\setlength{\tabcolsep}{5.5pt}
\renewcommand{\arraystretch}{1.12}
\caption{%
  \textbf{Main results on our controlled dial benchmark.}
  We report accuracy (\%) under clean, viewpoint-shifted (View), illumination-shifted (Illum), and combined conditions. Best results are highlighted in \textbf{bold}.
}
\label{tab:main_controlled}
\begin{tabular}{l cccc cccc}
\toprule
\multirow{2}{*}{\textbf{Model}} &
\multicolumn{4}{c}{\textbf{Clock}} &
\multicolumn{4}{c}{\textbf{Gauge}} \\
\cmidrule(lr){2-5} \cmidrule(l){6-9}
& Clean & View & Illum & Combined & Clean & View & Illum & Combined \\
\midrule
Molmo2-8B                 & 20.1 & 14.6 & 16.3 & 16.1 & 9.1  & 8.5  & 8.2  & 7.4  \\
Qwen3-VL-4B-Instruct      & 5.1  & 2.1  & 1.7  & 2.0  & 1.7  & 1.4  & 1.1  & 1.0  \\
Qwen3-VL-235B             & 5.4  & 2.8  & 3.4  & 2.7  & 3.2  & 2.3  & 2.5  & 2.1  \\
Gemini-3.1-Pro            & 6.3  & 4.3  & 4.6  & 3.1  & 13.8 & 11.2 & 10.7 & 8.7  \\
\midrule
\textbf{Qwen3-VL-4B-Instruct + TriSCA} & \textbf{37.4} & \textbf{28.7} & \textbf{26.6} & \textbf{20.0} & \textbf{26.2} & \textbf{22.5} & \textbf{20.6} & \textbf{18.4} \\
\bottomrule
\end{tabular}
\end{table*}
\subsection{Stage 2: Metadata-Grounded Observation-to-State Reasoning}
\label{subsec:stage2}

Stage~1 improves the geometry of the visual feature space, but reliable dial readout also requires the model to convert visual evidence into an explicit state estimate through a grounded observation-to-state process. Generic MLLMs often bypass this structure and jump directly from the image to a final answer, making them vulnerable to appearance-level shortcuts.

We therefore continue training with metadata-grounded supervised fine-tuning. Each training sample contains an image together with a structured target response that decomposes dial readout into interpretable intermediate steps---identifying the relevant indicator, estimating its dial-relative position, mapping it onto the calibrated dial scale, and producing the target state~\cite{yang2025can}. During training, we mask out the prompt tokens and apply supervision only to the grounded assistant response:
\[
\mathcal{L}_{\text{stage2}}=\mathcal{L}_{\text{sft}}.
\]

This supervision is shared across dial tasks through the same observation-to-state template. For clocks, the target state is instantiated as a time reading; for gauges, it is a scalar dial value. What changes across tasks is the calibration rule, not the reasoning structure. The role of Stage~2 is therefore not merely to improve text generation quality, but to transfer the state-aware bias introduced in Stage~1 into the decoding process, encouraging the model to ground its final answer in explicit dial evidence rather than superficial global appearance.

\subsection{Stage 3: State-Aware Objective Alignment}
\label{subsec:stage3}

Even with improved representation and grounded reasoning, the learning signal should still match the structure of the task. Exact-match supervision is too coarse for dial-based readout: a prediction that is slightly off from the correct dial state is clearly better than one that is far away, yet both are treated equally as incorrect by a discrete reward.

To address this, we introduce a state-aware reward defined in the canonical dial state space. Let $\hat{s}$ and $s$ denote the predicted and ground-truth dial states, respectively:
\[
r_{\text{state}}
=
\exp\!\left(
-\frac{d_{\text{state}}(\hat{s},s)^2}{2\sigma^2}
\right),
\]
where $d_{\text{state}}(\hat{s},s)$ measures the discrepancy between predicted and target dial states, and $\sigma$ controls the tolerance of the reward. For clocks, $d_{\text{state}}$ is instantiated using the circular discrepancy between predicted and target hand configurations; for gauges, it uses the normalized scalar error or corresponding pointer-angle discrepancy under gauge calibration.

We additionally include a lightweight format reward $r_{\text{fmt}}$ to encourage valid structured outputs, yielding the final reward:
\[
r=(1-\beta)r_{\text{state}}+\beta r_{\text{fmt}}.
\]

To optimize this objective efficiently, we employ Group Relative Policy Optimization (GRPO)~\cite{shao2024deepseekmath}. Unlike standard ~\cite{schulman2017proximal}, GRPO eliminates the need for a separate critic model by sampling a group of candidate outputs for the same input and normalizing their rewards within the group. This design is particularly suitable for multimodal large models, as it aligns the policy directly with continuous dial-state rewards while keeping optimization lightweight and stable.

\subsection{Stage-wise Optimization Summary}
\label{subsec:overall_framework}

Our training pipeline is deliberately stage-wise rather than relying on a single monolithic objective. Stage~1 first reshapes the visual feature space through state-aware representation alignment, improving same-state compactness and neighboring-state discriminability. Stage~2 then performs metadata-grounded observation-to-state supervision, encouraging the model to decode dial states through an explicit grounded readout process. Finally, Stage~3 replaces coarse discrete correctness with state-aware continuous rewards, so that the optimization signal better reflects physical state proximity.

The three stages are complementary but not equally positioned. Stage~1 addresses representational misalignment, Stage~2 reduces reasoning shortcuts, and Stage~3 refines the optimization objective once the first two stages have established a sufficiently strong foundation. Together, they progressively align representation, reasoning, and objective design with the intrinsic geometry of dial-based measurement reading.

\begin{table*}[t]
\centering
\small
\setlength{\tabcolsep}{4.5pt}
\renewcommand{\arraystretch}{1.12}
\caption{
\textbf{Comparison on MeasureBench.}
We report representative closed-source and open-source MLLMs from MeasureBench~\cite{lin2025measurebench} together with our additional evaluations.
Val and Unit columns are omitted for compactness.
\textbf{Ovr} denotes overall accuracy (\%); \textbf{Dial}, \textbf{Dig}, \textbf{Lin}, and \textbf{Com} denote category-wise accuracy (\%) on dial, digital, line, and complex measurement subsets.
Best results highlighted in \textbf{bold} within each source group.
}
\label{tab:measurebench_full}
\resizebox{0.8\textwidth}{!}{
\begin{tabular}{l ccccc ccccc}
\toprule
\small
\multirow{2}{*}{\textbf{Model}} &
\multicolumn{5}{c}{\textbf{Real-world subset}} &
\multicolumn{5}{c}{\textbf{Synthetic subset}} \\
\cmidrule(lr){2-6} \cmidrule(lr){7-11}
& \textbf{Ovr} & \textbf{Dial} & \textbf{Dig} & \textbf{Lin} & \textbf{Com}
& \textbf{Ovr} & \textbf{Dial} & \textbf{Dig} & \textbf{Lin} & \textbf{Com} \\
\midrule
\multicolumn{11}{l}{\textbf{Closed-source models}} \\
\midrule
Gemini-2.5-Pro   & \textbf{30.2} & \textbf{31.5} & \textbf{80.2} & \textbf{21.9} & \textbf{3.8} & \textbf{26.3} & \textbf{18.3} & 70.0 & \textbf{40.0} & \textbf{15.0} \\
Gemini-2.5-Flash & 20.2 & 20.5 & 65.6 & 13.0 & 1.0 & 18.1 & 11.9 & \textbf{75.0} & 25.7 & 1.7 \\
GPT-5            & 19.8 & 18.3 & 66.7 & 15.2 & 2.9 & 16.9 & 9.7  & 48.3 & 31.7 & 1.7 \\
GPT-5-Mini       & 22.0 & 20.8 & 70.8 & 16.9 & 2.9 & 17.9 & 12.0 & 56.7 & 28.3 & 1.7 \\
\midrule
\multicolumn{11}{l}{\textbf{Open-source models}} \\
\midrule
Qwen3-VL-8B                            & 15.3 & 14.5 & 53.1 & 11.3 & 0.0 & 11.4 & 8.0  & 25.0 & 19.3 & 0.0 \\
Qwen2.5-VL-72B                         & 14.5 & 12.2 & \textbf{55.2} & \textbf{12.2} & 0.0 & 11.7 & 6.4  & \textbf{43.3} & 21.0 & 0.0 \\
Qwen2.5-VL-32B                         & 11.7 & 9.0  & 51.6 & 9.7  & 0.0 & 10.5 & 5.3  & 28.3 & 22.0 & 0.0 \\
Qwen2.5-VL-7B                          & 14.6 & 13.8 & 49.0 & 11.4 & 0.0 & 10.9 & 5.7  & 33.3 & \textbf{21.7} & 0.0 \\
Qwen3-VL-4B-Instruct                   & 13.7 & 12.0 & 43.8 & 11.9 & 1.0 & 10.6 & 10.1 & 11.7 & 21.3 & 0.0 \\
\textbf{Qwen3-VL-4B-Instruct + TriSCA} & \textbf{20.5} & \textbf{28.5} & 41.7 & 10.4 & \textbf{1.9} & \textbf{13.5} & \textbf{13.1} & 10.0 & 21.0 & 0.0 \\
\bottomrule
\end{tabular}
}
\end{table*}

\begin{table}[t]
\centering
\small
\setlength{\tabcolsep}{4pt}
\renewcommand{\arraystretch}{1.1}
\caption{
\textbf{Continuous-state evaluation on the controlled clock benchmark.}
We report Tolerance-1 (Tol-1), Tolerance-5 (Tol-5), and MAE (minutes) under Clean and Combined settings.
Best results in \textbf{bold}.
}
\label{tab:clock_continuous_metrics}
\begin{tabular}{lcccc}
\toprule
\textbf{Model} & \textbf{Tol-1} & \textbf{Tol-5} & \textbf{MAE} $\downarrow$ \\
\midrule
\multicolumn{4}{l}{\textit{Clean}} \\
\midrule
Molmo2-8B                         & 39.4 & 63.6 & 32.31 \\
Qwen3-VL-4B-Instruct              & 16.8 & 40.4 & 53.96 \\
Gemini-3.1-Pro                    & 22.0 & 41.0 & 63.12 \\
\textbf{Qwen3-VL-4B + TriSCA}     & \textbf{67.2} & \textbf{77.6} & \textbf{23.54} \\
\midrule
\multicolumn{4}{l}{\textit{Combined}} \\
\midrule
Molmo2-8B                       & 18.0 & 29.8 & 81.25 \\
Qwen3-VL-4B-Instruct            & 8.2  & 17.2 & 99.77 \\
Gemini-3.1-Pro                  & 13.6 & 25.4 & 72.93 \\
\textbf{Qwen3-VL-4B + TriSCA}   & \textbf{45.6} & \textbf{58.0} & \textbf{43.62} \\
\bottomrule
\end{tabular}
\end{table}





\begin{table}[t]
\centering
\small
\setlength{\tabcolsep}{5.5pt}
\renewcommand{\arraystretch}{1.1}
\caption{
\textbf{Ablation study on the Combined split.}
Accuracy (\%) under simultaneous viewpoint and illumination perturbations.
}
\label{tab:ablation_combined}
\begin{tabular}{ccc cc}
\toprule
\textbf{Stage 1} & \textbf{Stage 2} & \textbf{Stage 3} & \textbf{Clock} & \textbf{Gauge} \\
\midrule
   &   &   & 2.0  & 1.0  \\
\checkmark &   &   & 7.0  & 5.3  \\
 & \checkmark &   & 5.0  & 3.2  \\
 &  & \checkmark & 1.8  & 1.0  \\
 & \checkmark & \checkmark & 5.6  & 3.0  \\
\checkmark &  & \checkmark & 7.2  & 5.8  \\
\checkmark & \checkmark &    & 11.0 & 12.1 \\
\checkmark & \checkmark & \checkmark & \textbf{20.0} & \textbf{18.4} \\
\bottomrule
\end{tabular}
\end{table}

\section{Experiments}
\label{sec:exp}

\subsection{Experimental Setup}
\label{subsec:exp_setup}

\paragraph{Implementation Details.}
In Stage~1, we freeze the language model and train only the visual tower for 3 epochs
(lr $= 2\times10^{-5}$, triplet loss weight warmed up to $0.1$).
In Stages~2 and~3, we perform full-parameter fine-tuning for 3 epochs
(lr $= 2\times10^{-7}$, cosine scheduler).
Our base model is Qwen3-VL-4B-Instruct.
Stage~1 uses 400K synthetic triplets; Stage~2 uses 200K grounded synthetic samples;
Stage~3 uses 1,273 real-world dial images collected from online sources.

\paragraph{Benchmarks.}
We evaluate on two benchmark types.
(1)~\textbf{Controlled benchmarks}: synthetic clock and gauge sets with four splits---
\textbf{Clean} (mild variation), \textbf{View} (camera-pose perturbation),
\textbf{Illum} (lighting/reflection/blur perturbation), and
\textbf{Combined} (both simultaneously, the hardest setting)---
further bucketed by perturbation severity for degradation analysis.
(2)~\textbf{MeasureBench}~\cite{lin2025measurebench}: an external real-world benchmark
for generalization assessment.
We report Exact Match as the primary metric, supplemented by
Tolerance-1/5 (Tol-1/5) and MAE (minutes) for clock readout.

\subsection{Main Results on Controlled Benchmarks}
\label{subsec:main_controlled}

Table~\ref{tab:main_controlled} shows that TriSCA substantially outperforms all baselines
across both dial types and all four conditions.
The largest relative gains occur on the Combined split,
consistent with our diagnosis that the primary bottleneck is
state-inconsistent representations under appearance variation rather than raw recognition difficulty.
Figure~\ref{fig:degradation_curve} further shows that TriSCA degrades substantially
more slowly as perturbation severity increases, confirming improved invariance to
state-preserving appearance changes.
\begin{figure*}[t]
    \centering
    \includegraphics[width=0.49\textwidth]{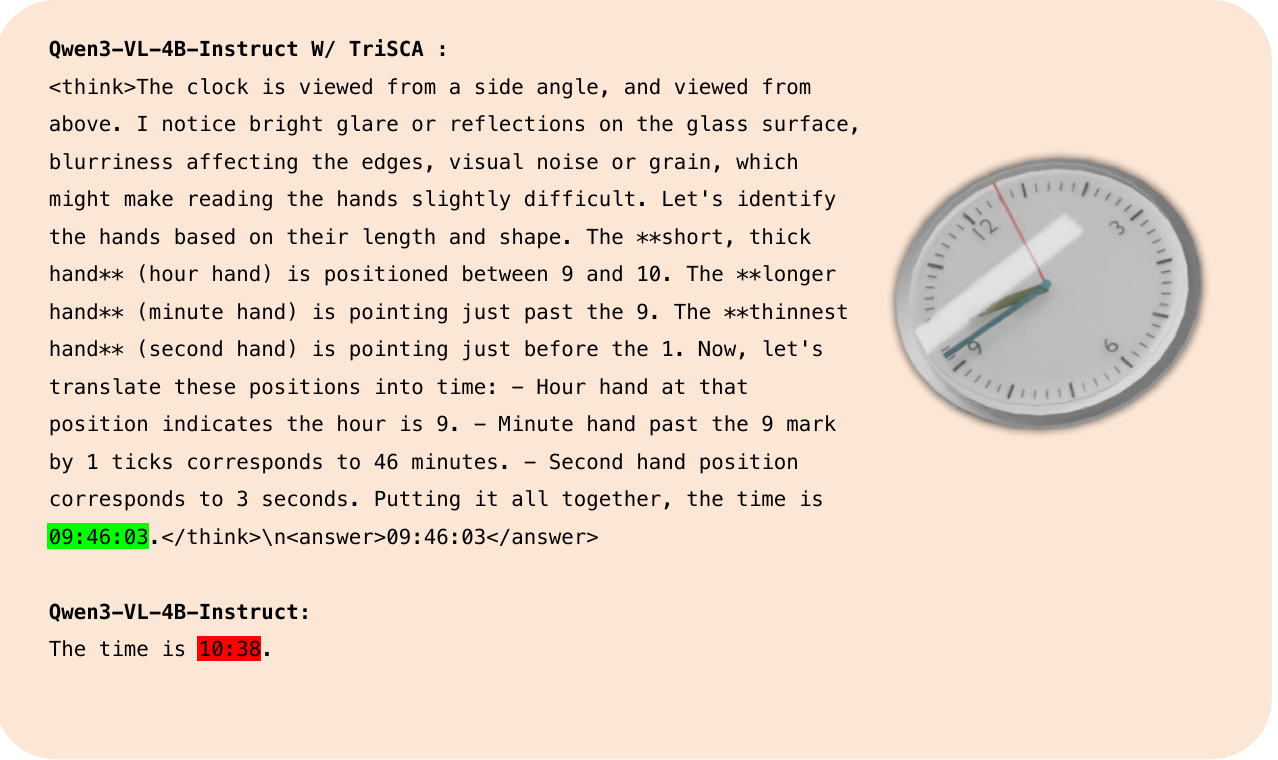}
    \hfill
    \includegraphics[width=0.49\textwidth]{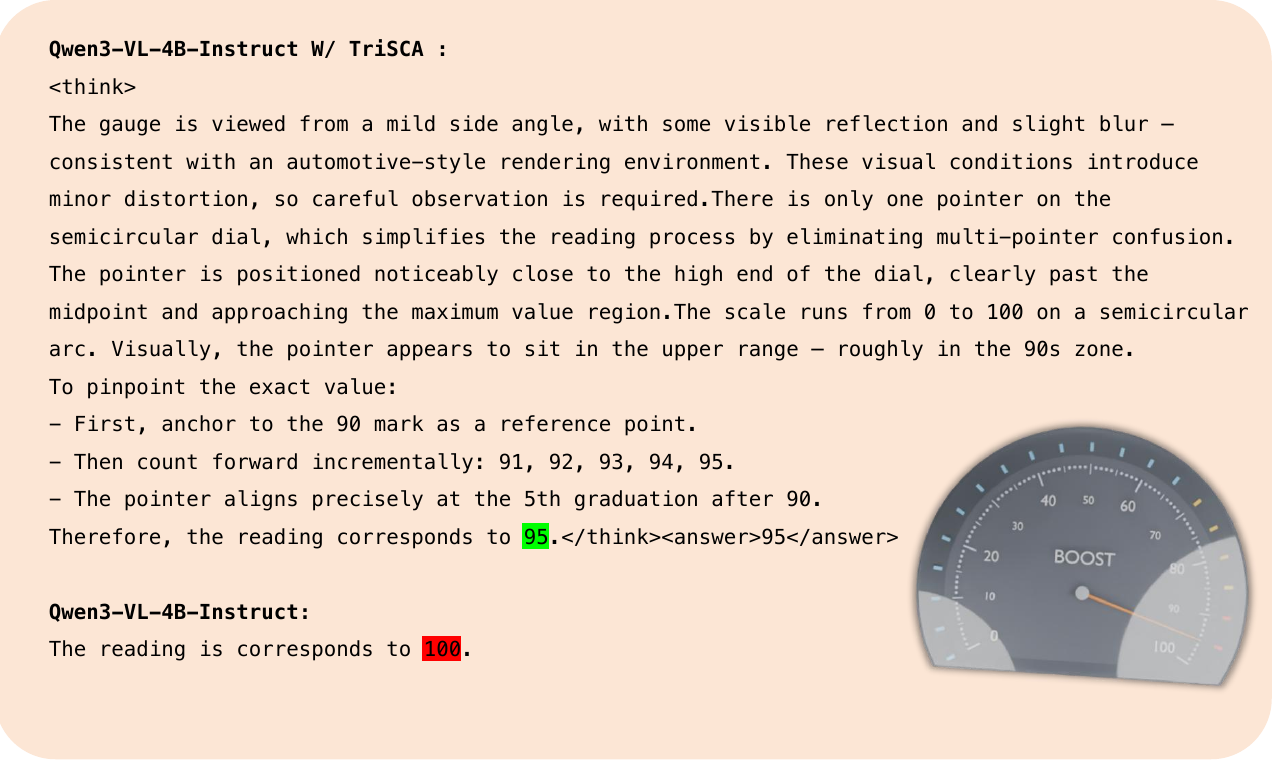}
    \caption{Representative case studies showing improved state consistency and local precision after alignment.}
    \label{fig:case_study}
\end{figure*}
\paragraph{Continuous-State Evaluation on Clock.}
Table~\ref{tab:clock_continuous_metrics} shows that TriSCA improves not only Exact Match
but also Tol-1, Tol-5, and MAE under both Clean and Combined settings.
The MAE reduction indicates that even failed exact-match predictions remain
far more tightly concentrated around the true dial state---a direct consequence
of the improved state geometry enforced by tri-level alignment.

\subsection{Generalization on MeasureBench}
\label{subsec:measurebench}

Table~\ref{tab:measurebench_full} reports results on MeasureBench~\cite{lin2025measurebench}.
Among open-source models, TriSCA achieves the largest gain on the Dial category
(+16.5 real-world, +3.0 synthetic over the Qwen3-VL-4B-Instruct base),
while performance on less related categories (Dig, Lin, Com) remains roughly stable.
This pattern confirms that our alignment is task-targeted rather than broadly disruptive,
and that the gains transfer to real-world dial images beyond our synthesis setting.

\subsection{Ablation Study}
\label{subsec:ablation}

Table~\ref{tab:ablation_combined} reports per-stage contributions on the Combined split.
Stage~1 is the strongest standalone component, confirming that representation alignment
is the primary driver. Stage~3 alone yields negligible gain, but contributes a
substantial additional +9.0/+6.3 points on top of Stage~1+2,
demonstrating that continuous state-aware rewards are effective only once
stable representations and grounded decoding are already in place.
The full three-stage pipeline achieves the best results on both dial types.

\subsection{Representation Analysis and Case Study}
\label{subsec:repr_analysis}

Figure~\ref{fig:tsne_compare} and Table~\ref{tab:retrieval_metrics} jointly confirm
that alignment reshapes the feature space toward dial-state consistency:
Recall@1 improves from 77.33 to 90.00 and Silhouette Score from 0.0725 to 0.3409.
Qualitative case studies in Figure~\ref{fig:case_study} further show that the
base model frequently drifts under appearance changes and confuses neighboring
states, whereas TriSCA produces more stable and locally precise predictions.

\section{Conclusion}
\label{sec:conclusion}

We present TriSCA, a tri-level state-consistent alignment framework that addresses
a fundamental representational bottleneck in dial-based measurement reading:
current MLLMs organize their feature space around superficial appearance rather
than the underlying physical state, causing brittle predictions under viewpoint
and illumination shifts. By aligning representations, reasoning, and training
objectives with the intrinsic geometry of the dial state space, TriSCA
consistently improves robustness on controlled clock and gauge benchmarks,
reduces continuous-state prediction error, and transfers to real-world evaluation
on MeasureBench---all while remaining task-targeted rather than broadly disruptive
to unrelated categories. Despite these gains, the current scope is limited to
dial-type instruments, and absolute performance under strong perturbations remains
far from satisfactory. A promising direction for future work is to extend
state-consistent alignment beyond dials to other instrument categories such as
linear scales and composite displays, as well as to more complex real-world
environments and broader forms of physically grounded structured estimation in MLLMs.

\bibliographystyle{ACM-Reference-Format}
\bibliography{references}

@article{wang2024qwen2,
  title={Qwen2-vl: Enhancing vision-language model's perception of the world at any resolution},
  author={Wang, Peng and Bai, Shuai and Tan, Sinan and Wang, Shijie and Fan, Zhihao and Bai, Jinze and Chen, Keqin and Liu, Xuejing and Wang, Jialin and Ge, Wenbin and others},
  journal={arXiv preprint arXiv:2409.12191},
  year={2024}
}

@inproceedings{deitke2025molmo,
  title={Molmo and pixmo: Open weights and open data for state-of-the-art vision-language models},
  author={Deitke, Matt and Clark, Christopher and Lee, Sangho and Tripathi, Rohun and Yang, Yue and Park, Jae Sung and Salehi, Mohammadreza and Muennighoff, Niklas and Lo, Kyle and Soldaini, Luca and others},
  booktitle={Proceedings of the Computer Vision and Pattern Recognition Conference},
  pages={91--104},
  year={2025}
}

@inproceedings{fu2024blink,
  title={Blink: Multimodal large language models can see but not perceive},
  author={Fu, Xingyu and Hu, Yushi and Li, Bangzheng and Feng, Yu and Wang, Haoyu and Lin, Xudong and Roth, Dan and Smith, Noah A and Ma, Wei-Chiu and Krishna, Ranjay},
  booktitle={European Conference on Computer Vision},
  pages={148--166},
  year={2024},
  organization={Springer}
}

@inproceedings{salomon2020deep,
  title={Deep learning for image-based automatic dial meter reading: Dataset and baselines},
  author={Salomon, Gabriel and Laroca, Rayson and Menotti, David},
  booktitle={2020 International joint conference on neural networks (IJCNN)},
  pages={1--8},
  year={2020},
  organization={IEEE}
}

@article{salomon2022image,
  title={Image-based automatic dial meter reading in unconstrained scenarios},
  author={Salomon, Gabriel and Laroca, Rayson and Menotti, David},
  journal={Measurement},
  volume={204},
  pages={112025},
  year={2022},
  publisher={Elsevier}
}

@inproceedings{reitsma2024under,
  title={Under pressure: learning-based analog gauge reading in the wild},
  author={Reitsma, Maurits and Keller, Julian and Blomqvist, Kenneth and Siegwart, Roland},
  booktitle={2024 IEEE International Conference on Robotics and Automation (ICRA)},
  pages={14--20},
  year={2024},
  organization={IEEE}
}

@article{saxena2025lost,
  title={Lost in time: Clock and calendar understanding challenges in multimodal LLMs},
  author={Saxena, Rohit and Gema, Aryo Pradipta and Minervini, Pasquale},
  journal={arXiv preprint arXiv:2502.05092},
  year={2025}
}

@article{shao2024deepseekmath,
  title={Deepseekmath: Pushing the limits of mathematical reasoning in open language models},
  author={Shao, Zhihong and Wang, Peiyi and Zhu, Qihao and Xu, Runxin and Song, Junxiao and Bi, Xiao and Zhang, Haowei and Zhang, Mingchuan and Li, YK and Wu, Yang and others},
  journal={arXiv preprint arXiv:2402.03300},
  year={2024}
}

@article{kamoi2024visonlyqa,
  title={Visonlyqa: Large vision language models still struggle with visual perception of geometric information},
  author={Kamoi, Ryo and Zhang, Yusen and Das, Sarkar Snigdha Sarathi and Zhang, Ranran Haoran and Zhang, Rui},
  journal={arXiv preprint arXiv:2412.00947},
  year={2024}
}

@inproceedings{kanade2025doyouseeme,
  title={Do you see me: A multidimensional benchmark for evaluating visual perception in multimodal llms},
  author={Kanade, Aditya Sanjiv and Ganu, Tanuja},
  booktitle={Proceedings of the 19th Conference of the European Chapter of the Association for Computational Linguistics (Volume 1: Long Papers)},
  pages={7285--7326},
  year={2026}
}

@article{stogiannidis2025mindthegap,
  title={Mind the gap: Benchmarking spatial reasoning in vision-language models},
  author={Stogiannidis, Ilias and McDonagh, Steven and Tsaftaris, Sotirios A},
  journal={arXiv preprint arXiv:2503.19707},
  year={2025}
}

@inproceedings{leonalcazar2024gauges,
  title={Learning to read analog gauges from synthetic data},
  author={Leon-Alcazar, Juan and Alnumay, Yazeed and Zheng, Cheng and Trigui, Hassane and Patel, Sahejad and Ghanem, Bernard},
  booktitle={Proceedings of the IEEE/CVF Winter Conference on Applications of Computer Vision},
  pages={8616--8625},
  year={2024}
}

@article{lin2025measurebench,
  title={Do Vision-Language Models Measure Up? Benchmarking Visual Measurement Reading with MeasureBench},
  author={Lin, Fenfen and Liu, Yesheng and Xu, Haiyu and Yue, Chen and He, Zheqi and Zhao, Mingxuan and Chen, Miguel Hu and Liu, Jiakang and Yao, JG and Yang, Xi},
  journal={arXiv preprint arXiv:2510.26865},
  year={2025}
}

@article{cheng2025visualthoughts,
  title={Visual thoughts: A unified perspective of understanding multimodal chain-of-thought},
  author={Cheng, Zihui and Chen, Qiguang and Xu, Xiao and Wang, Jiaqi and Wang, Weiyun and Fei, Hao and Wang, Yidong and Wang, Alex Jinpeng and Chen, Zhi and Che, Wanxiang and others},
  journal={arXiv preprint arXiv:2505.15510},
  year={2025}
}

@inproceedings{he2026finer,
  title={Fine-R1: Make Multi-modal LLMs Excel in Fine-Grained Visual Recognition by Chain-of-Thought Reasoning},
  author={He, Hulingxiao and Geng, Zijun and Peng, Yuxin},
  booktitle={The Fourteenth International Conference on Learning Representations}
}

@inproceedings{yang2022s,
  title={It's about time: Analog clock reading in the wild},
  author={Yang, Charig and Xie, Weidi and Zisserman, Andrew},
  booktitle={Proceedings of the IEEE/CVF Conference on Computer Vision and Pattern Recognition},
  pages={2508--2517},
  year={2022}
}

@article{fu2025have,
  title={Have Multimodal Large Language Models (MLLMs) Really Learned to Tell the Time on Analog Clocks?},
  author={Fu, Tairan and Gonz{\'a}lez, Miguel and Conde, Javier and Merino-G{\'o}mez, Elena and Reviriego, Pedro},
  journal={IEEE Internet Computing},
  year={2025},
  publisher={IEEE}
}

@article{choi2026s,
  title={It's Time to Get It Right: Improving Analog Clock Reading and Clock-Hand Spatial Reasoning in Vision-Language Models},
  author={Choi, Jaeha and Lee, Jin Won and You, Siwoo and Lee, Jangho},
  journal={arXiv preprint arXiv:2603.08011},
  year={2026}
}

@article{zhang2023multimodal,
  title={Multimodal chain-of-thought reasoning in language models},
  author={Zhang, Zhuosheng and Zhang, Aston and Li, Mu and Zhao, Hai and Karypis, George and Smola, Alex},
  journal={arXiv preprint arXiv:2302.00923},
  year={2023}
}

@inproceedings{schroff2015facenet,
  title={Facenet: A unified embedding for face recognition and clustering},
  author={Schroff, Florian and Kalenichenko, Dmitry and Philbin, James},
  booktitle={Proceedings of the IEEE conference on computer vision and pattern recognition},
  pages={815--823},
  year={2015}
}

@article{khosla2020supervised,
  title={Supervised contrastive learning},
  author={Khosla, Prannay and Teterwak, Piotr and Wang, Chen and Sarna, Aaron and Tian, Yonglong and Isola, Phillip and Maschinot, Aaron and Liu, Ce and Krishnan, Dilip},
  journal={Advances in neural information processing systems},
  volume={33},
  pages={18661--18673},
  year={2020}
}

@article{comanici2025gemini,
  title={Gemini 2.5: Pushing the frontier with advanced reasoning, multimodality, long context, and next generation agentic capabilities},
  author={Comanici, Gheorghe and Bieber, Eric and Schaekermann, Mike and Pasupat, Ice and Sachdeva, Noveen and Dhillon, Inderjit and Blistein, Marcel and Ram, Ori and Zhang, Dan and Rosen, Evan and others},
  journal={arXiv preprint arXiv:2507.06261},
  year={2025}
}

@inproceedings{yue2024mmmu,
  title={Mmmu: A massive multi-discipline multimodal understanding and reasoning benchmark for expert agi},
  author={Yue, Xiang and Ni, Yuansheng and Zhang, Kai and Zheng, Tianyu and Liu, Ruoqi and Zhang, Ge and Stevens, Samuel and Jiang, Dongfu and Ren, Weiming and Sun, Yuxuan and others},
  booktitle={Proceedings of the IEEE/CVF conference on computer vision and pattern recognition},
  pages={9556--9567},
  year={2024}
}

@inproceedings{yue-etal-2025-mmmu,
    title = "{MMMU}-Pro: A More Robust Multi-discipline Multimodal Understanding Benchmark",
    author = "Yue, Xiang  and
      Zheng, Tianyu  and
      Ni, Yuansheng  and
      Wang, Yubo  and
      Zhang, Kai  and
      Tong, Shengbang  and
      Sun, Yuxuan  and
      Yu, Botao  and
      Zhang, Ge  and
      Sun, Huan  and
      Su, Yu  and
      Chen, Wenhu  and
      Neubig, Graham",
    editor = "Che, Wanxiang  and
      Nabende, Joyce  and
      Shutova, Ekaterina  and
      Pilehvar, Mohammad Taher",
    booktitle = "Proceedings of the 63rd Annual Meeting of the Association for Computational Linguistics (Volume 1: Long Papers)",
    month = jul,
    year = "2025",
    address = "Vienna, Austria",
    publisher = "Association for Computational Linguistics",
    url = "https://aclanthology.org/2025.acl-long.736/",
    doi = "10.18653/v1/2025.acl-long.736",
    pages = "15134--15186",
    ISBN = "979-8-89176-251-0"
}

@inproceedings{singh2019towards,
  title={Towards vqa models that can read},
  author={Singh, Amanpreet and Natarajan, Vivek and Shah, Meet and Jiang, Yu and Chen, Xinlei and Batra, Dhruv and Parikh, Devi and Rohrbach, Marcus},
  booktitle={Proceedings of the IEEE/CVF conference on computer vision and pattern recognition},
  pages={8317--8326},
  year={2019}
}

@article{liu2024ocrbench,
  title={Ocrbench: on the hidden mystery of ocr in large multimodal models},
  author={Liu, Yuliang and Li, Zhang and Huang, Mingxin and Yang, Biao and Yu, Wenwen and Li, Chunyuan and Yin, Xu-Cheng and Liu, Cheng-Lin and Jin, Lianwen and Bai, Xiang},
  journal={Science China Information Sciences},
  volume={67},
  number={12},
  pages={220102},
  year={2024},
  publisher={Springer}
}

@article{cheng2025v,
  title={V-star: Benchmarking video-llms on video spatio-temporal reasoning},
  author={Cheng, Zixu and Hu, Jian and Liu, Ziquan and Si, Chenyang and Li, Wei and Gong, Shaogang},
  journal={arXiv preprint arXiv:2503.11495},
  year={2025}
}

@article{hu2025tinyalign,
  title={TinyAlign: Boosting Lightweight Vision-Language Models by Mitigating Modal Alignment Bottlenecks},
  author={Hu, Yuanze and Fan, Zhaoxin and Wang, Xinyu and Li, Gen and Qiu, Ye and Yang, Zhichao and Wu, Wenjun and Wu, Kejian and Sun, Yifan and Deng, Xiaotie and others},
  journal={arXiv preprint arXiv:2505.12884},
  year={2025}
}

@article{schulman2017proximal,
  title={Proximal policy optimization algorithms},
  author={Schulman, John and Wolski, Filip and Dhariwal, Prafulla and Radford, Alec and Klimov, Oleg},
  journal={arXiv preprint arXiv:1707.06347},
  year={2017}
}

@article{bai2025qwen3,
  title={Qwen3-vl technical report},
  author={Bai, Shuai and Cai, Yuxuan and Chen, Ruizhe and Chen, Keqin and Chen, Xionghui and Cheng, Zesen and Deng, Lianghao and Ding, Wei and Gao, Chang and Ge, Chunjiang and others},
  journal={arXiv preprint arXiv:2511.21631},
  year={2025}
}

@article{yang2025can,
  title={Can Structured Templates Facilitate LLMs in Tackling Harder Tasks?: An Exploration of Scaling Laws by Difficulty},
  author={Yang, Zhichao and Fan, Zhaoxin and Li, Gen and Hu, Yuanze and Wang, Xinyu and Qiu, Ye and Wang, Xin and Sun, Yifan and Wu, Wenjun},
  journal={arXiv preprint arXiv:2508.19069},
  year={2025}
}

@misc{vicuna,
    title = {Vicuna: An Open-Source Chatbot Impressing GPT-4 with 90\%* ChatGPT Quality},
    url = {https://lmsys.org/blog/2023-03-30-vicuna/},
    author = {Chiang, Wei-Lin and Li, Zhuohan and Lin, Zi and Sheng, Ying and Wu, Zhanghao and Zhang, Hao and Zheng, Lianmin and Zhuang, Siyuan and Zhuang, Yonghao and Gonzalez, Joseph E. and Stoica, Ion and Xing, Eric P.},
    month = {March},
    year = {2023}
}

@misc{cogvlm,
    title={CogVLM: Visual Expert for Pretrained Language Models}, 
    author={Weihan Wang and Qingsong Lv and Wenmeng Yu and Wenyi Hong and Ji Qi and Yan Wang and Junhui Ji and Zhuoyi Yang and Lei Zhao and Xixuan Song and Jiazheng Xu and Bin Xu and Juanzi Li and Yuxiao Dong and Ming Ding and Jie Tang},
    year={2023},
    eprint={2311.03079},
    archivePrefix={arXiv},
    primaryClass={cs.CV},
    url={https://arxiv.org/abs/2311.03079}
}

@misc{moe-llava,
    title={MoE-LLaVA: Mixture of Experts for Large Vision-Language Models}, 
    author={Bin Lin and Zhenyu Tang and Yang Ye and Jinfa Huang and Junwu Zhang and Yatian Pang and Peng Jin and Munan Ning and Jiebo Luo and Li Yuan},
    year={2024},
    eprint={2401.15947},
    archivePrefix={arXiv},
    primaryClass={cs.CV},
    url={https://arxiv.org/abs/2401.15947}
}

@misc{shikra,
    title={Shikra: Unleashing Multimodal LLM's Referential Dialogue Magic}, 
    author={Keqin Chen and Zhao Zhang and Weili Zeng and Richong Zhang and Feng Zhu and Rui Zhao},
    year={2023},
    eprint={2306.15195},
    archivePrefix={arXiv},
    primaryClass={cs.CV},
    url={https://arxiv.org/abs/2306.15195}
}

@misc{hill,
    title={HILL: A Hallucination Identifier for Large Language Models}, 
    author={Florian Leiser and Sven Eckhardt and Valentin Leuthe and Merlin Knaeble and Alexander Maedche and Gerhard Schwabe and Ali Sunyaev},
    year={2024},
    eprint={2403.06710},
    archivePrefix={arXiv},
    primaryClass={cs.HC},
    url={https://arxiv.org/abs/2403.06710}
}

@misc{ascd,
    title={ASCD: Attention-Steerable Contrastive Decoding for Reducing Hallucination in MLLM}, 
    author={Yujun Wang and Aniri and Jinhe Bi and Soeren Pirk and Yunpu Ma},
    year={2025},
    eprint={2506.14766},
    archivePrefix={arXiv},
    primaryClass={cs.CV},
    url={https://arxiv.org/abs/2506.14766}
}

@misc{param_hall,
    title={Detecting and Preventing Hallucinations in Large Vision Language Models}, 
    author={Anisha Gunjal and Jihan Yin and Erhan Bas},
    year={2023},
    eprint={2308.06394},
    archivePrefix={arXiv},
    primaryClass={cs.CV},
    url={https://arxiv.org/abs/2308.06394}
}

@misc{param_hall1,
    title={Mitigating Object Hallucinations in Large Vision-Language Models through Visual Contrastive Decoding}, 
    author={Sicong Leng and Hang Zhang and Guanzheng Chen and Xin Li and Shijian Lu and Chunyan Miao and Lidong Bing},
    year={2023},
    eprint={2311.16922},
    archivePrefix={arXiv},
    primaryClass={cs.CV},
    url={https://arxiv.org/abs/2311.16922}
}

@misc{minigpt4,
    title={MiniGPT-4: Enhancing Vision-Language Understanding with Advanced Large Language Models}, 
    author={Deyao Zhu and Jun Chen and Xiaoqian Shen and Xiang Li and Mohamed Elhoseiny},
    year={2023},
    eprint={2304.10592},
    archivePrefix={arXiv},
    primaryClass={cs.CV},
    url={https://arxiv.org/abs/2304.10592}
}

@misc{sharegpt4v,
    title={ShareGPT4V: Improving Large Multi-Modal Models with Better Captions}, 
    author={Lin Chen and Jinsong Li and Xiaoyi Dong and Pan Zhang and Conghui He and Jiaqi Wang and Feng Zhao and Dahua Lin},
    year={2023},
    eprint={2311.12793},
    archivePrefix={arXiv},
    primaryClass={cs.CV},
    url={https://arxiv.org/abs/2311.12793}
}

@misc{hall_tune,
    title={Mitigating Hallucination in Large Multi-Modal Models via Robust Instruction Tuning}, 
    author={Fuxiao Liu and Kevin Lin and Linjie Li and Jianfeng Wang and Yaser Yacoob and Lijuan Wang},
    year={2023},
    eprint={2306.14565},
    archivePrefix={arXiv},
    primaryClass={cs.CV},
    url={https://arxiv.org/abs/2306.14565}
}

@misc{dola,
    title={DoLa: Decoding by Contrasting Layers Improves Factuality in Large Language Models}, 
    author={Yung-Sung Chuang and Yujia Xie and Hongyin Luo and Yoon Kim and James Glass and Pengcheng He},
    year={2023},
    eprint={2309.03883},
    archivePrefix={arXiv},
    primaryClass={cs.CL},
    url={https://arxiv.org/abs/2309.03883}
}
\end{document}